\documentclass[conference]{IEEEtran}
\IEEEoverridecommandlockouts
\usepackage{cite}
\usepackage{amsmath,amssymb,amsfonts}
\usepackage{graphicx}
\usepackage{textcomp}
\usepackage{xcolor}
\usepackage{bm, bbm}
\usepackage{algorithm, algorithmic}
\usepackage{multirow}
\usepackage[caption=false]{subfig}
\usepackage{url}

\newtheorem{definition}{Definition}[section]

\renewcommand{\algorithmicrequire}{\textbf{Input: }}
\renewcommand{\algorithmicensure}{\textbf{Output: }}

\def\BibTeX{{\rm B\kern-.05em{\sc i\kern-.025em b}\kern-.08em
    T\kern-.1667em\lower.7ex\hbox{E}\kern-.125emX}}

\begin{document}

\title{On Prediction Feature Assignment in the Heckman Selection Model}

\author{\IEEEauthorblockN{Huy Mai, Xintao Wu}
\IEEEauthorblockA{\textit{Department of Electrical Engineering and Computer Science} \\
\textit{University of Arkansas}\\
Fayetteville, Arkansas, USA \\
\{huymai, xintaowu\}@uark.edu}}

\maketitle

\begin{abstract}
Under missing-not-at-random (MNAR) sample selection bias, the performance of a prediction model is often degraded. This paper focuses on one classic instance of MNAR sample selection bias where a subset of samples have non-randomly missing outcomes. The Heckman selection model and its variants have commonly been used to handle this type of sample selection bias. The Heckman model uses two separate equations to model the prediction and selection of samples, where the selection features include all prediction features. When using the Heckman model, the prediction features must be properly chosen from the set of selection features. However, choosing the proper prediction features is a challenging task for the Heckman model. This is especially the case when the number of selection features is large. Existing approaches that use the Heckman model often provide a manually chosen set of prediction features. In this paper, we propose Heckman-FA as a novel data-driven framework for obtaining prediction features for the Heckman model. Heckman-FA first trains an assignment function that determines whether or not a selection feature is assigned as a prediction feature. Using the parameters of the trained function, the framework extracts a suitable set of prediction features based on the goodness-of-fit of the prediction model given the chosen prediction features and the correlation between noise terms of the prediction and selection equations. Experimental results on real-world datasets show that Heckman-FA produces a robust regression model under MNAR sample selection bias.
\end{abstract}

\begin{IEEEkeywords}
sample selection bias, missing-not-at-random, Heckman selection model, robust regression
\end{IEEEkeywords}

\section{Introduction}
Regression is sensitive to dataset shift \cite{moreno2012unifying}, where the training and testing sets come from different distributions. Dataset shift can arise due to sample selection bias, where a sample is non-uniformly chosen from a population for training a model. This type of bias can cause a subset of training samples to be partially observed, where any of the covariates or outcome of a sample is missing, or completely unobserved. Consequently, the performance of the model after training on this biased set will be degraded when the model is deployed. Most approaches such as \cite{chen2016robust}, \cite{lei2021near}, and \cite{sahoo2022learning} handle the missing-at-random (MAR) setting, where the selection of training samples is assumed to be independent from the outcome given the covariates. However, these approaches cannot properly account for the missing-not-at-random (MNAR) setting, where the selection of training samples is assumed to not be independent from the outcome given the covariates. 


In this work, we focus on the problem of MNAR sample selection bias on the outcome. As a motivating example, consider the relationship between SAT score (feature) and the amount of scholarship offered by a certain university (outcome), where some students have missing values of scholarship. There could be some hidden mechanism behind the missing outcomes. For instance, amounts of scholarship offered to students who have not declared their majors are not collected. When the undeclared students are omitted from the training, a biased model is produced and could be very different from the ground truth model that would have been trained had scholarship amounts of all students been collected. However, under MNAR sample selection on the outcome, we leverage the observed feature information of records with missing outcomes to train a model that is close to the ground truth model.

The Heckman selection model \footnote{In our paper, we use 'Heckman selection model' and 'Heckman model' interchangeably.} \cite{heckman1979sample} is a Nobel Prize winning approach that addresses MNAR sample selection bias on the outcome. The method involves two equations: the prediction equation and the selection equation. The prediction equation describes the relationship between the covariates and the outcome of interest. The selection equation specifies the probability that a sample is selected in the training data. Estimation of the selection equation requires an exclusion restriction, where the selection equation includes one or more variables that are covariates. To handle MNAR sample selection bias on the outcome, the Heckman model considers the respective noise terms of the prediction and selection equations, which follow a bivariate normal distribution.

Although the presence of an exclusion restriction avoids multicollinearity for the prediction model \cite{leung1996choice}, the process to identify a valid exclusion restriction is often difficult in practice. This is first due to the lack of clear theoretical knowledge on which selection features should be included in the prediction model \cite{genback2015uncertainty}. Moreover, using the Heckman selection model with an invalid exclusion restriction can lead to non-robust regression on the biased training set \cite{wolfolds2019misaccounting}. Choosing from features that affect the selection, one way to address these challenges is to search through all combinations of selection features to find a suitable set of prediction features for the Heckman selection model. However, this search process becomes computationally expensive as we deal with a large number of selection features in real-world data.

\subsection{Problem Formulation}
We provide a list of important symbols used throughout the paper in Table \ref{tab:notation} in the Appendix. In our work, we generally use unbolded letters to denote scalars, bold lower-case letters to denote vectors, and bold upper-case letters to denote matrices. For accented letters, we use a hat to denote estimations, tilde to denote approximations, and underline to denote augmented vectors or matrices. As exceptions to these conventions, we use $\bm{X}_{\cdot k}$ to denote the $k$th column of any matrix and $\bm{\pi}$ to denote the probability matrix. 

We let $\mathcal{X}$ be the feature space and $\mathcal{Y}$ be the continuous target attribute. We also denote $\mathcal{D}_{tr} = \{\bm{t}_i\}_{i=1}^n$ as the training set of $n$ samples that are originally sampled from the population to be modeled yet biased under MNAR sample selection bias. We define each sample $\bm{t}_i$ as
\begin{equation} \small
    \bm{t}_i = \begin{cases}
    (\bm{x}_i,y_i,s_i=1) & 1\leq i\leq m \\
    (\bm{x}_i, s_i=0) & m+1\leq i\leq n
    \end{cases}
\end{equation} 
where the binary variable $s_i$ indicates whether or not $y_i$ is observed. We define $\mathcal{D}_s$ as the set containing the first $m$ training samples where each sample is fully observed and $\mathcal{D}_u$ as the set that contains the remaining $n-m$ training samples with unobserved outcome. 

\begin{definition}[MNAR Sample Selection]
     Suppose that prediction features $\bm{x}^{(p)}_i\subset \bm{x}_i$ are used to describe $y_i$. Missing-not-at-random (MNAR) sample selection occurs for a sample $\bm{t}_i$ if $s_i$ is not independent of $y_i$ given $\bm{x}^{(p)}_i$, i.e. $P(s_i\vert \bm{x}^{(p)}_i, y_i) \neq P(s_i\vert \bm{x}^{(p)}_i)$.
\end{definition}

In other words, it is not sufficient to use prediction features $\bm{x}^{(p)}_i$ to describe MNAR sample selection on the outcome. To account for this selection mechanism, the Heckman model assumes that there exist selection features $\bm{x}^{(s)}_i$---all prediction features $\bm{x}^{(p)}_i$ and additional features that do not affect the outcome---that help model the selection mechanism. Choosing from all observed features $\bm{x}_i$, $\bm{x}^{(s)}_i$ can be specified by domain users or simply learned via goodness-of-fit. In this work, we assume that $\bm{x}_i=\bm{x}^{(s)}_i$ for simplicity.

\noindent \textbf{Problem Statement.} To perform regression, a linear model $h(\bm{x}_{i}^{(p)}; \bm{\beta})$ with prediction features $\bm{x}^{(p)}_i$ and parameters $\bm{\beta}$ is fitted by learning to minimize a loss function over $\mathcal{D}_{tr}$. Given that $\mathcal{D}_{tr}$ is biased due to MNAR sample selection, we seek to choose prediction features from a set of selection features using an assignment function $\psi$. Based on the extracted assignment of prediction features, a model $h(\underline{\bm{x}}^{(p)}_i;\underline{\bm{\beta}})$ is fitted after running the Heckman model, where $\underline{\bm{x}}^{(p)}_i$ denotes the extracted prediction features augmented with the inverse Mills ratio (IMR) and $\underline{\bm{\beta}}$ denotes the Heckman coefficients.

\subsection{Contributions}

In this work, we present the Heckman selection model with Feature Assignment (Heckman-FA) as a framework that finds a suitable assignment of prediction features for the Heckman model to robustly handle MNAR sample selection bias on the outcome. The core contributions of our work are summarized as follows. First, Heckman-FA trains an assignment function that determines whether a selection feature is assigned as a prediction feature or not. The assignment function is defined in terms of samples from the Gumbel-Softmax distribution \cite{jang2016categorical}. Second, using the parameters of the trained assignment function, Heckman-FA extracts prediction features for the Heckman model based on goodness-of-fit and the correlation between the noise terms of the prediction and selection equations. Third, we apply our method to real-world datasets and compare the performance of Heckman-FA against other regression baselines. We empirically show that Heckman-FA produces a robust regression model under MNAR sample selection bias and outperforms these regression baselines.

\section{Related Work}

\subsection{Incorporating the Heckman Selection Model}

The Heckman selection model has been widely utilized to handle MNAR sample selection bias in different fields such as criminology \cite{bushway2007magic} and epidemiology \cite{barnighausen2011correcting}. Variants of the Heckman selection model have also been proposed (see a comprehensive survey \cite{vella1998estimating}). In the area of fair machine learning, \cite{du2022fair} applied the Heckman model to correct MNAR sample selection bias while achieving fairness for linear regression models. Very recently, \cite{kahng2023domain} extended the Heckman model to model multiple domain-specific sample selection mechanisms and proposed to jointly learn prediction and domain selection to achieve generalization on the true population. For approaches that incorporate the Heckman model and its variants, the prediction features are manually chosen beforehand. Our work makes a data-driven choice for the prediction features after training a feature assignment function. 

Empirical analysis has often been used to examine the effect of exclusion restrictions on the performance of the Heckman model. \cite{leung1996choice}  conducted Monte Carlo experiments with and without exclusion restrictions and found that the Heckman model is susceptible to collinearity issues in the absence of exclusion restrictions. Recently, \cite{wolfolds2019misaccounting} conducted a simulation showing that ordinary least squares (OLS) on the biased training set outperforms the Heckman model based approaches that do not have a valid exclusion restriction.

For the Heckman model, the correlation between the two noise terms carries information about the sample selection process. Because the noise terms are unobserved, the true value of the correlation is unknown. \cite{genback2015uncertainty} used a range of values for the correlation to derive uncertainty intervals for the parameters of the prediction model. In our work, we also consider a range of correlation values. However, rather than defining the range of correlation values for a fixed set of prediction features, we specify the range of correlation values as we dynamically assign prediction features for the Heckman model. 

The idea of variable assignment based on reparametrized random samples has also been explored in previous work. For instance, \cite{lippe2022citris} proposed to learn an assignment function that maps each latent variable to some causal factor as part of learning a causal representation. Similar to our work, the assignment function is learned by sampling an assignment from the Gumbel-Softmax distribution. In our work, however, we do not map variables to causal factors. Instead, we map each variable to a value that indicates whether or not the variable is assigned as a feature for the predictive model.

\subsection{Learning under Biased Training Data}

Machine learning on missing training data has been well-studied. There are a number of techniques that handle MAR sample selection bias in the training set. Importance weighting \cite{shimodaira2000improving} is commonly used to handle the MAR setting to reweigh the training samples. However, it can result in inaccurate estimates due to the influence of data instances with large importance weights on the reweighted loss. To address this drawback, recent MAR approaches such as \cite{chen2016robust} have been constructed based on distributionally robust optimization of the reweighted loss. \cite{sahoo2022learning} introduced Rockafellar-Uryasev (RU) regression to produce a model that is robust against bounded MAR sample selection bias, where the level of distribution shift between the training and testing sets is restricted. Based on the assumption for MNAR sample selection bias, methods that account for MAR bias would not properly model the MNAR data mechanism. Thus we expect methods that handle MAR bias to not be robust against MNAR sample selection bias. On the other hand, our approach uses the Heckman selection model to model the MNAR data mechanism, where additional features that do not affect the outcome are needed to fit the selection mechanism. 



In particular, there are recent approaches that address the problem of MNAR labels in training data. In recommender learning, \cite{wang2019doubly} and \cite{lee2021dual} use separate propensity estimation models to predict the observation of a label due to MNAR ratings and user feedback, respectively. Unlike our work, they consider matrix factorization as the prediction model, which is not for linear regression on tabular data.


In semi-supervised learning, \cite{hu2022non} employed class-aware propensity score and imputation strategies on the biased training set toward developing a model that is doubly robust against MNAR data. We emphasize that our problem setting is different than semi-supervised learning. In semi-supervised learning, unlabeled samples are separated into clusters based on similarities. However, in our problem setting, we do not perform clustering on the samples with missing labels.

\section{Heckman Selection Model Revisited}

Formally, the Heckman selection model \cite{heckman1979sample} models the selection and prediction equations as follows. The selection equation of the $i$th sample is 
\begin{equation} \small
d_i = \bm{x}_{i}^{(s)}\bm{\gamma}+u_{i}^{(s)} 
\end{equation}
 where $\bm{\gamma}$ is the set of regression coefficients for selection, $\bm{x}_{i}^{(s)}$ denotes the features for sample selection, and $u_{i}^{(s)}\sim \mathcal{N}(0,1)$ is the noise term for the selection equation. The selection value of the $i$th sample $s_i$ is defined as:
\begin{equation} \label{eq:selectioneq} \small
s_i =\left\{
\begin{aligned}
&1 \hspace{0.5cm} d_i > 0 \\
&0  \hspace{0.5cm} d_i \le 0 \\
\end{aligned}
\right.
\end{equation} 
The model learns the selection based on Eq. (\ref{eq:selectioneq}) and the prediction of the $i$th sample based on linear regression, with
\begin{equation} \small
    y_i = \bm{x}_i^{(p)}\bm{\beta} + u_i^{(p)}
\end{equation}
Assuming $u_i^{(p)}\sim \mathcal{N}(0, \sigma^{2})$, we define $u_i^{(p)}=\sigma \epsilon_i$ where $\epsilon_i\sim \mathcal{N}(0,1)$. Moreover, $u_i^{(p)}$ and $u_i^{(s)}$ are correlated with a correlation coefficient of $\rho$. If $\rho$ differs from zero, there is an indication that the missing observations are MNAR.

To correct the bias in $\mathcal{D}_{tr}$, the conditional expectation of the predicted outcome
\begin{equation} \label{eq:heckman} \small
    \begin{split} 
        \mathbb{E}[y_i\vert s_i=1] &= \mathbb{E}[y_i\vert d_i\geq 0] \\
        &= \mathbb{E}[\bm{x}_i^{(p)}\bm{\beta} + u_i^{(p)}\vert \bm{x}_{i}^{(s)}\bm{\gamma}+u_{i}^{(s)}> 0] \\
        &= \bm{x}_i^{(p)}\bm{\beta} + \mathbb{E}[u_i^{(p)}\vert \bm{x}_{i}^{(s)}\bm{\gamma}+u_{i}^{(s)}> 0] \\
        &= \bm{x}_i^{(p)}\bm{\beta} + \mathbb{E}[u_i^{(p)}\vert u_{i}^{(s)}> -\bm{x}_{i}^{(s)}\bm{\gamma}]
    \end{split}
\end{equation}
is computed over all samples in $\mathcal{D}_s$. Because $u_i^{(p)}\sim \mathcal{N}(0, \sigma^{2})$ and $u_i^{(s)}\sim \mathcal{N}(0, 1)$ are correlated,
\begin{equation} \small
    \mathbb{E}[u_i^{(p)}\vert u_{i}^{(s)}> -\bm{x}_{i}^{(s)}\bm{\gamma}] = \lambda_i\rho\sigma
\end{equation}
where $\lambda_i=\frac{\phi(-\bm{x}_{i}^{(s)}\bm{\gamma})}{1 - \Phi(-\bm{x}_{i}^{(s)}\bm{\gamma})}=\frac{\phi(-\bm{x}_{i}^{(s)}\bm{\gamma})}{\Phi(\bm{x}_{i}^{(s)}\bm{\gamma})}$ is the inverse Mills ratio (IMR). We denote $\phi(\cdot)$ as the standard normal density function and $\Phi(\cdot)$ as the standard normal cumulative distribution function. Thus Eq. (\ref{eq:heckman}) is rewritten as
\begin{equation} \label{eq:heckmanrewritten} \small
    \mathbb{E}[y_i\vert s_i=1] = \bm{x}_i^{(p)}\bm{\beta} + \lambda_i\rho\sigma = \underline{\bm{x}}^{(p)}_i\underline{\bm{\beta}}
\end{equation}
where $\underline{\bm{x}}^{(p)}_i = [\bm{x}^{(p)}_i, \lambda_i]$, $\underline{\bm{\beta}}=[\bm{\beta}, \beta_H]$, and $\beta_H=\rho\sigma$. 

Algorithm \ref{alg:heckman} gives pseudocode of executing the Heckman model, which follows two steps:

\noindent \textbf{Step 1.} Since there is no prior knowledge of the true value of $\lambda_i$ for each sample in $\mathcal{D}_s$, $\lambda_i$ is estimated as $\hat{\lambda}_i$ by first computing $\hat{\bm{\gamma}}$ using probit regression over $\mathcal{D}_{tr}$. As indicated in line 2 of Algorithm \ref{alg:heckman}, we estimate $\hat{\bm{\gamma}}$ after maximizing
\begin{equation} \label{eq:heckmangamma} \small
    \begin{split}
        \mathcal{L}(\bm{\gamma})   &= \prod_{i=1}^n P(s_i=1)^{s_i}\cdot P(s_i=0)^{1-s_i} \\
            &= \prod_{i=1}^n \Phi(\bm{x}^{(s)}_i\bm{\gamma})^{s_i} (1 - \Phi(\bm{x}^{(s)}_i\bm{\gamma}))^{1-s_i}
    \end{split}
\end{equation}
over $\mathcal{D}_{tr}$, where $\mathcal{L}(\cdot)$ is the likelihood. As shown in line 4, $\hat{\bm{\gamma}}$ is then used to compute $\hat{\lambda}_i$ for each $\bm{t}_i$ in $\mathcal{D}_s$.

\noindent \textbf{Step 2.} Using $\hat{\lambda}_i$, the prediction model is
\begin{equation} \small
    \begin{split}
        \hat{y}_i &= \bm{x}_i^{(p)}\bm{\beta} + \hat{\lambda}_i\beta_H
    \end{split}
\end{equation}
The estimated set of coefficients $\underline{\hat{\bm{\beta}}}$ is computed by minimizing $\sum_{i=1}^m (y_i - \hat{y}_i)^2$ over $\mathcal{D}_s$. As a result, as indicated in line 7, $\underline{\hat{\bm{\beta}}}$ is computed using the closed-form solution
\begin{equation} \label{eq:heckmanclosedform} \small
    \underline{\hat{\bm{\beta}}} = (\underline{\bm{X}}^{(p)T} \underline{\bm{X}}^{(p)})^{-1} \underline{\bm{X}}^{(p)T}\bm{y}
\end{equation}

\begin{algorithm}[t]
\caption{\textit{Heckman}}\label{alg:heckman}
\begin{flushleft}
    \algorithmicrequire $\,$ Training set $\mathcal{D}_{tr}=\{(\bm{x}^{(s)}_i, y_i, s_i=1)\}_{i=1}^{m}\cup \{(\bm{x}^{(s)}_i, s_i=0)\}_{i=m+1}^{n}$, selection feature matrix $\bm{X}^{(s)}$, prediction feature matrix $\bm{X}^{(p)}$ \\
    \algorithmicensure Estimated Heckman coefficients $\underline{\hat{\bm{\beta}}}$, IMR $\{\hat{\lambda}_i\}_{i=1}^m$
\end{flushleft}
\begin{algorithmic}[1]
    \STATE $\mathcal{D}_s\gets \{(\bm{x}^{(s)}_i, y_i, s_i=1)\}_{i=1}^{m}$
    \STATE Estimate $\hat{\bm{\gamma}}$ after maximizing Eq. (\ref{eq:heckmangamma}) on $\mathcal{D}_{tr}$ using $\bm{X}^{(s)}$
    \FOR{$\bm{t}_i\in \mathcal{D}_s$}
        \STATE $\hat{\lambda}_i\gets \frac{\phi(-\bm{x}_{i}^{(s)}\hat{\bm{\gamma}})}{\Phi(\bm{x}_{i}^{(s)}\hat{\bm{\gamma}})}$
        \STATE $\underline{\bm{x}}_i^{(p)}\gets [\bm{x}_i^{(p)}, \hat{\lambda}_i]$
    \ENDFOR
    \STATE Obtain $\underline{\hat{\bm{\beta}}}$ by evaluating Eq. (\ref{eq:heckmanclosedform}) on $\mathcal{D}_s$ using $\underline{\bm{X}}^{(p)}$
    \RETURN{$\underline{\hat{\bm{\beta}}}$ and $\{\hat{\lambda}_i\}_{i=1}^m$}
\end{algorithmic}
\end{algorithm}

\noindent \textbf{Exclusion Restriction.} The Heckman selection model assumes that the selection features consist of every prediction feature and additional features that do not affect the outcome. The selection and prediction features are generally not identical as it can introduce multicollinearity to the prediction model. Specifically, if the selection and prediction features are the same, the IMR $\lambda_i$ would be expressed as
\begin{equation} \small
    \begin{split}
        \lambda_i &= \frac{\phi(-\bm{x}_{i}^{(s)}\bm{\gamma})}{\Phi(\bm{x}_{i}^{(s)}\bm{\gamma})} = \frac{\phi(-\bm{x}_{i}^{(p)}\bm{\gamma})}{\Phi(\bm{x}_{i}^{(p)}\bm{\gamma})}
    \end{split}
\end{equation}
This would mean that Eq. (\ref{eq:heckmanrewritten}) is properly identified through the nonlinearity of $\lambda_i$. However, $\lambda_i$ is roughly linear over a wide range of values for $\bm{x}^{(s)}_i\bm{\gamma}$ \cite{puhani2000heckman}. Hence, Step 2 of the Heckman model would yield unrobust estimates for $\underline{\hat{\bm{\beta}}}$ due to the multicollinearity between $\bm{x}^{(p)}_i$ and $\lambda_i$.

\section{Methodology}

When choosing prediction features from a set of $K$ selection features, we encounter the following challenges. First, there are $2^K - 1$ possible choices to make for the set of prediction features. For any dataset that has a large number of selection features, searching for a suitable set of prediction features becomes computationally expensive. Second, the Heckman selection model does not perform well for exclusion restrictions that are not valid. In other words, some choices for the set of prediction features are not helpful when using the Heckman  model to handle MNAR sample selection bias on the outcome. 

We introduce Heckman-FA, a framework for using the Heckman selection model via a learned feature assignment. Heckman-FA first learns the weights of an assignment function $\psi$ that draws samples from the Gumbel-Softmax distribution \cite{jang2016categorical}. This function outputs an assignment of prediction features given a matrix $\bm{\pi}$ of probabilities of including a selection feature in the prediction model. The framework then uses this assignment of prediction features to run the Heckman  model and compute the mean absolute error (MAE) of predictions on $\mathcal{D}_s$ when using the Heckman selection model. To optimize $\psi$, we minimize the MAE with respect to $\bm{\pi}$. This results in an estimated probability matrix $\hat{\bm{\pi}}$. To extract the prediction features, Heckman-FA looks through different prediction feature matrices by sampling from the Gumbel-Softmax distribution using $\hat{\bm{\pi}}$. When determining the extracted prediction feature matrix, we first consider whether or not the estimated correlation between the noise terms is within a range that is user-defined based on prior domain knowledge. We further consider goodness-of-fit to ensure that the prediction model is of quality. Using the selection and extracted prediction features, we run the Heckman model to fit a robust prediction model under MNAR sample selection bias on the outcome.

\subsection{Assignment Function}

We formally introduce an assignment function $\psi(k)$, which is defined as
\begin{equation} \small
    \begin{split}
        \psi(k) &= \begin{cases}
            1 & k \text{th selection feature is assigned} \\
            0 & k \text{th selection feature is not assigned}
        \end{cases}
    \end{split}
\end{equation}
In general, an assignment function determines which of the $K$ selection features are also prediction features. 

\noindent \textbf{Assignment Computation.} In Algorithm \ref{alg:assign}, we provide the pseudocode for computing a matrix $\bm{X}^{(p)}$ of assigned prediction features from the selection feature matrix $\bm{X}^{(s)}$. We assume that both $\bm{X}^{(p)}$ and $\bm{X}^{(s)}$ have $K$ columns. However, we define the $J$ features assigned for prediction as the columns in $\bm{X}^{(p)}$ not equal to the zero vector $\bm{0}$.

To obtain $\bm{X}^{(p)}$ based on selection features, we compute the assignment $\psi(k)$ for the $k$th selection feature by drawing samples from a categorical distribution. Let $\bm{Z}\in \mathbb{R}^{2\times K}$ be a categorical sample matrix such that each element is either 0 or 1. The Gumbel-Max trick is used to efficiently draw categorical samples. Following the steps of \cite{jang2016categorical}, a sample $\bm{Z}_{\cdot k}$, the $k$th column of $\bm{Z}$, is drawn from a categorical distribution with class probabilities $\pi_{1k}$ and $\pi_{2k}$, where $\pi_{2k}$ ($\pi_{1k}$) is the probability that the $k$th selection feature is (not) assigned to the prediction model. $\bm{Z}_{\cdot k}$ is expressed as
\begin{equation} \label{eq:gumbelonehot} \small
    \bm{Z}_{\cdot k} = \text{one\_hot}(\underset{q}{\text{argmax}}[g_{qk} + \log \pi_{qk}])
\end{equation}
where $g_{qk}\sim \text{ Gumbel}(0,1)$ and $q\in \{1,2\}$. Thus we express the assignment $\psi$ for the $k$th selection feature as
\begin{equation}\label{eq:psi} \small
    \begin{split}
        \psi(k;\pi) &= \begin{bmatrix}
            0 \\
            1 \\
        \end{bmatrix}^T \bm{Z}_{\cdot k}
    \end{split}
\end{equation}
and compute $x_{ik}^{(p)} = x^{(s)}_{ik}\odot \psi(k;\bm{\pi})$ for each selection feature as indicated in lines 5 and 6 of Algorithm \ref{alg:assign}, using $\odot$ to denote elementwise multiplication.

\begin{algorithm}[t]
\caption{\textit{FeatureAssign}}\label{alg:assign}
\begin{flushleft}
    \algorithmicrequire $\,$ Training set $\mathcal{D}_{tr}=\{(\bm{x}^{(s)}_i, y_i, s_i=1)\}_{i=1}^{m}\cup \{(\bm{x}^{(s)}_i, s_i=0)\}_{i=m+1}^{n}$, number of selection features $K$, probability matrix $\bm{\pi}$ \\
    \algorithmicensure Prediction feature matrix $\bm{X}^{(p)}$
\end{flushleft}
\begin{algorithmic}[1]
    \STATE $\mathcal{D}_s\gets \{(\bm{x}^{(s)}_i, y_i, s_i=1)\}_{i=1}^{m}$
    \STATE Compute $\bm{Z}$ using Eq. (\ref{eq:gumbelonehot}) based on $\bm{\pi}$
    \FORALL{$\bm{t}_i\in \mathcal{D}_s$}
        \FOR{$k\in \{1,\ldots,K\}$}
            \STATE $\psi(k;\bm{\pi})\gets \begin{bmatrix}
                0 \\
                1 \\
            \end{bmatrix}^T \bm{Z}_{\cdot k}$
            \STATE $x_{ik}^{(p)}\gets x^{(s)}_{ik}\odot \psi(k;\bm{\pi})$
        \ENDFOR
    \ENDFOR
    \RETURN{$\bm{X}^{(p)}$}
\end{algorithmic}
\end{algorithm}

\noindent \textbf{Backpropagation.} To train the assignment function $\psi$, we learn $\hat{\pi}_{1k}$ and $\hat{\pi}_{2k}$ for each selection feature. However, $\bm{Z}_{\cdot k}$ is expressed in terms of argmax, which is non-differentiable. Thus we cannot derive $\nabla_{\bm{\pi}} \bm{Z}$ in order to learn $\hat{\bm{\pi}}$. On the other hand, we can use the Gumbel-Softmax distribution \cite{jang2016categorical} to approximate the categorical sample $\bm{Z}_{\cdot k}$. By the Straight-Through Gumbel Estimator, we compute $\tilde{\bm{Z}}\in \mathbb{R}^{2\times K}$ as a continuous, differentiable approximation of argmax, where
\begin{equation} \label{eq:continuousapprox} \small
    \tilde{z}_{qk} = \frac{\text{exp}((\log \pi_{qk} + g_{qk})/\tau)}{\sum_{r=1}^{2} \text{exp}((\log \pi_{rk} + g_{rk})/\tau)}
\end{equation}
with $\tau$ as the softmax temperature. Notice that as $\tau\rightarrow 0$, Eq. (\ref{eq:continuousapprox}) will approximate the argmax function in Eq. (\ref{eq:gumbelonehot}), and the Gumbel-Softmax sample vector will approach a one-hot vector. Hence $\nabla_{\bm{\pi}} \bm{Z}\approx \nabla_{\bm{\pi}} \tilde{\bm{Z}}$, where
\begin{equation} \label{eq:tildezpartial} \small
    \begin{split}
        \frac{ \partial\tilde{z}_{qk} }{ \partial \pi_{qk} } &= \frac{\prod_{r=1}^2\text{exp}((\log \pi_{rk} + g_{rk})/\tau)}{(\sum_{r=1}^{2} \text{exp}((\log \pi_{rk} + g_{rk})/\tau))^2} \cdot \frac{1}{\tau\pi_{qk}} 
    \end{split}
\end{equation}
Therefore, although we use Eq. (\ref{eq:psi}) to obtain assignments, we use $\tilde{\bm{Z}}$ instead of $\bm{Z}$ when performing backpropagation.

Based on Eq. (\ref{eq:tildezpartial}), $\nabla_{\bm{\pi}} \tilde{\bm{Z}}$ is well-defined, so we are able to train the assignment function and estimate parameters $\hat{\bm{\pi}}$. Formally, we compute a probability matrix $\hat{\bm{\pi}}$ such that
\begin{equation}\label{eq:optim} \small
    \begin{split}
        \hat{\bm{\pi}} &=  \underset{\bm{\pi}}{\text{argmin}}\, \mathcal{L}_{MAE} \\ 
            &= \underset{\bm{\pi}}{\text{argmin}}\left( \frac{1}{m} \sum_{i=1}^m \bigg\vert y_i - (\bm{x}_i^{(p)}\bm{\beta} + \hat{\lambda}_i\beta_H) \bigg\vert \right)
    \end{split}
\end{equation}
where $x_{ik}^{(p)} = x^{(s)}_{ik}\odot \psi(k;\bm{\pi})$ for the $k$th selection feature. To ensure that the Heckman model is a quality prediction model, we consider the predictive performance of the Heckman model on $\mathcal{D}_s$ to learn the assignment function. In this work, we choose to minimize the MAE over $\mathcal{D}_s$ in order to obtain $\hat{\bm{\pi}}$. For $\nabla_{\bm{\pi}} \mathcal{L}_{MAE}$, using on Eq. (\ref{eq:tildezpartial}), we see that
\begin{equation} \label{eq:lmaegrad} \small
    \begin{split}
        \frac{ \partial \mathcal{L}_{MAE} }{ \partial \pi_{qk} } &= \frac{ \partial \mathcal{L}_{MAE} }{ \partial\hat{y}_i } \cdot \frac{ \partial\hat{y}_i }{ \partial x^{(p)}_{ik} } \cdot \frac{ \partial x^{(p)}_{ik} }{ \partial \psi(k) } \cdot \frac{ \partial \psi(k) }{ \partial\tilde{z}_{qk} }\cdot \frac{ \partial\tilde{z}_{qk} }{ \partial \pi_{qk} } \\
            &=  -\frac{1}{m} \sum_{i=1}^m \frac{ y_i - \hat{y}_i}{\vert y_i - \hat{y}_i \vert} \cdot \beta_{k} \cdot x^{(s)}_{ik} \cdot \frac{ \partial \psi(k) }{ \partial\tilde{z}_{qk} } \\
            &\,\, \cdot \frac{\prod_{r=1}^2\text{exp}((\log \pi_{rk} + g_{rk})/\tau)}{(\sum_{r=1}^{2} \text{exp}((\log \pi_{rk} + g_{rk})/\tau))^2} \cdot \frac{1}{\tau\pi_{qk}}
    \end{split}
\end{equation}
where $\displaystyle \frac{ \partial \psi(k) }{ \partial\tilde{z}_{2k} } = 1$. Thus $\nabla_{\bm{\pi}}\mathcal{L}_{MAE}$ is well-defined. 


We note that when training the assignment function, we don't minimize other metrics such as mean squared error (MSE) and root mean squared error (RMSE). If these metrics are minimized as an objective to obtain $\hat{\bm{\pi}}$, then the assignment function parameters do not change at all during backpropagation. A proof of this claim is provided in the Appendix.


\subsection{Extraction of Suitable Assignment} \label{sec:extraction}

After we train the assignment function $\psi$, we utilize the estimated parameters $\hat{\bm{\pi}}$ to extract a suitable set of features for the prediction model. We propose a sampling-based strategy to extract $\bm{X}^{(p)}$ using $\hat{\bm{\pi}}$. We base this strategy on the estimated correlation $\hat{\rho}$ between the noise terms $u^{(s)}_i$ and $u^{(p)}_i$ and the adjusted $R^2$ value, denoted as $R^2_a$. Specifically, a suitable prediction feature matrix is extracted based on a user-defined range of values for the correlation. The range of values are provided based on prior domain knowledge on a given dataset. Moreover, we observe that multiple prediction feature assignments can correspond to a correlation that is within the user-defined range. To further decide on which prediction feature matrix to extract, we also consider goodness-of-fit to ensure that the prediction model is of quality.

In general, $\rho$ carries information about the nature of the sample selection process. However, because $u^{(s)}_i$ and $u^{(p)}_i$ are unobserved for all $\bm{t}_i$, the true value of $\rho$ is unknown for a dataset. We must instead consider the estimated correlation $\hat{\rho}$. To compute $\hat{\rho}$, we note that a consistent estimator of $\sigma^{2}$ can be derived based on the Heckman selection model. First, define $v_i = y_i - \hat{y}_i$ as the error of predicting the outcome of the $i$th sample using the Heckman selection model. The true conditional variance of $v_i$ is
\begin{equation} \small
    \begin{split}
        \mathbb{E}[v^{2}_i\vert s_i=1] &= \sigma^{2}( 1 + \rho^2( \lambda_i(-\bm{x}^{(s)}_i\bm{\gamma}) - \lambda^2_i ) ) \\
            &= \sigma^{2} + (\rho\sigma)^2( \lambda_i(-\bm{x}^{(s)}_i\bm{\gamma}) - \lambda^2_i )
    \end{split}
\end{equation}
Consider the average conditional variance over all $\bm{t}_i\in\mathcal{D}_s$, where
\begin{equation} \label{eq:plim} \small
    \begin{split}
        \text{plim }\frac{1}{m}\sum_{i=1}^m \mathbb{E}[v^{2}_i\vert s_i=1] &= \sigma^{2}\left( 1 + \frac{\rho^2}{m}\sum_{i=1}^m \lambda_i(-\bm{x}^{(s)}_i\gamma) - \lambda^2_i  \right) \\
            &= \sigma^{2} + \frac{(\rho\sigma)^2}{m}\sum_{i=1}^m \lambda_i(-\bm{x}^{(s)}_i\gamma) - \lambda^2_i
    \end{split}
\end{equation}
with plim denoting convergence in probability. The average conditional variance over all $\bm{t}_i\in\mathcal{D}_s$ is estimated using MSE
\begin{equation} \label{eq:heckmanmse} \small
    \begin{split} 
        \frac{1}{m}\sum_{i=1}^m v^{2}_i &= \frac{1}{m}\sum_{i=1}^m (y_i - \hat{y}_i)^2
    \end{split}
\end{equation}
of predicting the outcome using the Heckman model. Thus, using Eq. (\ref{eq:plim}) and (\ref{eq:heckmanmse}), $\sigma^{2}$ can be estimated as
\begin{equation} \label{eq:sigmasq} \small
    \begin{split}
        \hat{\sigma}^{2} &= \frac{1}{m} \sum_{i=1}^m (y_i - \hat{y}_i)^2 - \frac{\hat{\beta}_H^2}{m} \sum_{i=1}^m \hat{\lambda}_i(-\bm{x}^{(s)}_i\hat{\bm{\gamma}}) - \hat{\lambda}^2_i
    \end{split}
\end{equation}
where $\hat{\beta}_H$ is an estimate of $\rho\sigma$. Therefore, using Eq. (\ref{eq:sigmasq}), we obtain $\displaystyle \hat{\rho} = \hat{\beta}_H /\hat{\sigma}$.

\begin{algorithm}[t]
\caption{\textit{Extraction}}\label{alg:evaloptimalassignment}
\begin{flushleft}
    \algorithmicrequire $\,$ Training set $\mathcal{D}_{tr}=\{(\bm{x}^{(s)}_i, y_i, s_i=1)\}_{i=1}^{m}\cup \{(\bm{x}^{(s)}_i, s_i=0)\}_{i=m+1}^{n}$, selection feature matrix $\bm{X}^{(s)}$, number of selection features $K$, estimated parameters $\hat{\bm{\pi}}$, correlation threshold $[\rho_{min}, \rho_{max}]$ number of Gumbel-Softmax samples $B$ \\
    \algorithmicensure Prediction feature matrix $\bm{X}^{(p)}$
\end{flushleft}
\begin{algorithmic}[1]
    \STATE $R^{2*}_a\gets -\infty$
    \FOR{$B$ iterations}
        \STATE $\bm{X}^{(p)}_{temp}\gets $ \textit{FeatureAssign}($\mathcal{D}_{tr}$, $K$, $\hat{\bm{\pi}}$)
        \STATE $\underline{\hat{\bm{\beta}}}, \{\hat{\lambda}_i\}_{i=1}^m\gets $ \textit{Heckman}($\mathcal{D}_{tr}$, $\bm{X}^{(s)}$, $\bm{X}^{(p)}_{temp}$)
        \STATE Compute $\hat{\sigma}^{2}$ using Eq. (\ref{eq:sigmasq})
        \STATE $\hat{\rho}\gets \hat{\beta}_H / \hat{\sigma}$
        \STATE Compute $R^2_a$ using Eq. (\ref{eq:adjustedr2})
        \IF{$\hat{\rho}\in [\rho_{min}, \rho_{max}]$ and $R^2_a > R^{2*}_a$}
            \STATE $\bm{X}^{(p)}\gets \bm{X}^{(p)}_{temp}$
            \STATE $R^{2*}_a\gets R^2_a$
        \ENDIF
    \ENDFOR
    \RETURN{$\bm{X}^{(p)}$}
\end{algorithmic}
\end{algorithm}

We consider a set of prediction features to be suitable for the Heckman model if $\hat{\rho}$ is within the range $[\rho_{min}, \rho_{max}]$, where the values of $\rho_{min}$ and $\rho_{max}$ are user-specified. We work with this user-specified range because the true value of $\rho$ is unknown. We expect that the values of $\rho_{min}$ and $\rho_{max}$ are appropriately chosen to indicate that the Heckman model properly handles MNAR sample selection bias. On one hand, the range $[\rho_{min}, \rho_{max}]$ should not contain 0 since $\rho=0$ indicates MAR sample selection bias. On the other hand, $\rho_{min}$ and $\rho_{max}$ should not be too negative or too positive since strong correlation renders the Heckman model unstable \cite{nawata1993note}.

Because there may be multiple values of $\hat{\rho}$ that are in $[\rho_{min}, \rho_{max}]$, we also obtain $R^2_a$ by computing
\begin{equation} \label{eq:adjustedr2} \small
    \begin{split}
        R^2_a &= 1 - \frac{ (1 - R^2) (m - 1) }{m - J - 1}
    \end{split}
\end{equation}
where $R^2$ is the coefficient of determination and $J$ is the number of prediction features. Having the largest number of prediction features such that $\hat{\rho}$ is in $[\rho_{min}, \rho_{max}]$ does not imply that the linear model is the best fitted. Thus $R^2_a$ is helpful since it also factors in the number of prediction features when measuring the goodness-of-fit of a model. 

As our strategy to extract $\bm{X}^{(p)}$ for the Heckman selection model, we propose to collect $B$ Gumbel-Softmax samples based on $\hat{\bm{\pi}}$. Out of the $B$ samples, we choose the prediction feature matrix such that the prediction model has the highest value of $R^2_a$ given that $\hat{\rho}$ is in $[\rho_{min}, \rho_{max}]$. The pseudocode for this process is listed in Algorithm \ref{alg:evaloptimalassignment}. In lines 2-12, we iteratively look for $\bm{X}^{(p)}$. In line 3, we sample $\bm{X}^{(p)}_{temp}$ by executing Algorithm \ref{alg:assign} based on $\hat{\bm{\pi}}$. In line 4, we execute the Heckman model based on $\underline{\bm{X}}^{(p)}_{temp}$ and get $\underline{\hat{\bm{\beta}}}$ and $\{\hat{\lambda}_i\}_{i=1}^m$. In line 5, we compute $\hat{\sigma}^{(p)2}$ using Eq. (\ref{eq:sigmasq}). In line 6, we calculate $\hat{\rho}$ by dividing $\hat{\beta}_H$ by $\hat{\sigma}$. In line 7, we compute $R^2_a$. Throughout each iteration, we check whether or not $\hat{\rho}$ is in $[\rho_{min},\rho_{max}]$ and $R^2_a$ is larger than the current largest $R^{2*}_a$ value. If this condition is satisfied, then we update $\bm{X}^{(p)}$, as indicated in line 9. In line 13, $\bm{X}^{(p)}$ is returned.

\begin{algorithm}[t]
\caption{Heckman-FA}\label{alg:heckmanpsi}
\begin{flushleft}
    \algorithmicrequire $\,$ Training set $\mathcal{D}_{tr}=\{(\bm{x}^{(s)}_i, y_i, s_i=1)\}_{i=1}^{m}\cup \{(\bm{x}^{(s)}_i, s_i=0)\}_{i=m+1}^{n}$, selection feature matrix $\bm{X}^{(s)}$, number of selection features $K$, initial fixed value $c$, number of training epochs $T$, learning rate $\alpha$, correlation threshold $[\rho_{min}, \rho_{max}]$ number of Gumbel-Softmax samples $B$ \\
    \algorithmicensure Augmented prediction feature matrix $\underline{\bm{X}}^{(p)}$, estimated Heckman coefficients $\underline{\hat{\bm{\beta}}}$
\end{flushleft}
\begin{algorithmic}[1]
    \FOR{$K$ selection features}
        \STATE Initialize $\pi_{2k} = c$ and $\pi_{1k} = 1 - c$
    \ENDFOR
    \FOR{$T$ epochs}
        \STATE $\bm{X}^{(p)}\gets $ \textit{FeatureAssign}($\mathcal{D}_{tr}$, $K$, $\bm{\pi}$)
        \STATE $\underline{\hat{\bm{\beta}}}, \{\hat{\lambda}_i\}_{i=1}^m\gets $ \textit{Heckman}($\mathcal{D}_{tr}$, $\bm{X}^{(s)}$, $\bm{X}^{(p)}$)
        \STATE Compute $\mathcal{L}_{MAE}$ using $\underline{\hat{\bm{\beta}}}$ and $\{\hat{\lambda}_i\}_{i=1}^m$
        \STATE $\bm{\pi}\gets \bm{\pi} - \displaystyle \alpha\nabla_{\bm{\pi}}\mathcal{L}_{MAE}$
        \STATE $\hat{\bm{\pi}}\gets \bm{\pi}$
    \ENDFOR
    \STATE $\bm{X}^{(p)}\gets $ \textit{Extraction}($\mathcal{D}_{tr}$, $\bm{X}^{(s)}$, $K$, $\hat{\bm{\pi}}$, $\rho_{min}$, $\rho_{max}$, $B$)
    \STATE $\underline{\hat{\bm{\beta}}}, \{\hat{\lambda}_i\}_{i=1}^m\gets $ \textit{Heckman}($\mathcal{D}_{tr}$, $\bm{X}^{(s)}$, $\bm{X}^{(p)}$)
    \STATE $\underline{\bm{X}}^{(p)} \gets [\bm{X}^{(p)}, \{\hat{\lambda}_i\}_{i=1}^m]$
    \RETURN{$\underline{\bm{X}}^{(p)}$ and $\underline{\hat{\bm{\beta}}}$}
\end{algorithmic}
\end{algorithm}

\begin{table*}[t]
    \centering
    \caption{Comparison of Heckman-FA against baselines using extracted prediction features.}
    \resizebox{0.85\textwidth}{!}{\begin{tabular}{|c|c|c||c|c||c|c||c|c|}
        \hline
        \multirow{2}{*}{Extracted prediction features} & \multirow{2}{*}{$\hat{\rho}$} & \multirow{2}{*}{$R^2_a$} & \multicolumn{2}{c||}{\textbf{Naive}} & \multicolumn{2}{c||}{\textbf{RU}\cite{sahoo2022learning}} & \multicolumn{2}{c|}{\textbf{Heckman-FA}} \\ \cline{4-9}
        & & & Train MSE & Test MSE & Train MSE & Test MSE & Train MSE & Test MSE \\
        \hline
        \hline
        \multicolumn{9}{|c|}{CRIME} \\
        \hline
        \hline
        1, 2, 3, 4, 5, 7, 8, & & & & & & & & \\
        11, 15, 16, 21, 22, 23, 24 & 0.0423 & 0.4780 & 0.0159 & 0.0218 & 0.0109 & 0.0212 & 0.0159 & \textbf{0.0203} \\
        \hline
        \hline
        \multicolumn{9}{|c|}{COMPAS} \\
        \hline
        \hline
        1, 2, 3, 5, 7, 10 & 0.3179 & 0.3454 & 0.0761 & 0.2568 & 0.0749 & 0.2531 & 0.0759 & \textbf{0.2504} \\
        \hline
    \end{tabular}}
    \label{tab:sampling}
\end{table*}

\begin{table}[t]
    \centering
    \caption{Testing MSEs of Heckman-FA using different values of $T$ and $c$.}
    \resizebox{0.29\textwidth}{!}{\begin{tabular}{|c||c|c|c|c|}
        \hline
        \multirow{2}{*}{$c$} & \multicolumn{4}{c|}{$T$ training epochs} \\ \cline{2-5}
         & 100 & 500 & 1,000 & 2,000 \\
        \hline
        \hline
        \multicolumn{5}{|c|}{CRIME} \\
        \hline
        \hline
        0.25 & 0.0205 & 0.0210 & 0.0209 & 0.0211 \\
        0.5 & 0.0211 & 0.0214 & 0.0211 & 0.0207 \\
        0.75 & 0.0217 & 0.0216 & 0.0212 & 0.0208 \\
        \hline
        \hline
        \multicolumn{5}{|c|}{COMPAS} \\
        \hline
        \hline
        0.25 & 0.2514 & 0.2506 & 0.2530 & 0.2507 \\
        0.5 & 0.2502 & 0.2505 & 0.2502 & 0.2502 \\
        0.75 & 0.2502 & 0.2502 & 0.2506 & 0.2502 \\
        \hline
    \end{tabular}}
    \label{tab:testingmsespi0}
\end{table}

\subsection{Heckman Selection Model with Feature Assignment}

Algorithm \ref{alg:heckmanpsi} gives the pseudocode of Heckman-FA. In lines 1-3, we initialize $\bm{\pi}$. In this work, we use a fixed value $c\in (0,1)$ to initialize $\pi_{2k}$ and $\pi_{1k} = 1 - \pi_{2k}$ for the $k$th selection feature. This means that each selection feature has an equal probability of being assigned as a prediction feature when we start training $\psi$. For $c$, we give users some flexibility to choose which value to use. However, we suggest for users to not use values that are extremely close to 0 or 1 for $c$. This ensures that $\psi$ is trained on a variety of sets of prediction features based on random Gumbel-Softmax samples. In lines 5-9, $\psi$ is trained over $T$ epochs. In line 5, we obtain the prediction feature matrix $\bm{X}^{(p)}$ by executing Algorithm \ref{alg:assign}. In line 6, we execute the steps of the Heckman model to get $\underline{\hat{\bm{\beta}}}$ and $\{\hat{\lambda}_i\}_{i=1}^m$. In line 7, we use $\underline{\hat{\bm{\beta}}}$ and $\{\hat{\lambda}_i\}_{i=1}^m$ to compute $\mathcal{L}_{MAE}$. In line 8, we update $\bm{\pi}$ by computing $\nabla_{\pi}\mathcal{L}_{MAE}$ using Eq. (\ref{eq:lmaegrad}). In line 11, using $\hat{\bm{\pi}}$, we extract $\bm{X}^{(p)}$ by calling Algorithm \ref{alg:evaloptimalassignment}. In line 12, we run the Heckman  model using the extracted prediction features and obtain $\underline{\hat{\bm{\beta}}}$ and $\{\hat{\lambda}_i\}_{i=1}^m$. Letting $\underline{\bm{X}}^{(p)}$ be the concatenation of $\bm{X}^{(p)}$ and $\{\hat{\lambda}_i\}_{i=1}^m$ in line 13, Heckman-FA returns $\underline{\bm{X}}^{(p)}$ and $\underline{\hat{\bm{\beta}}}$. 

\noindent \textbf{Heckman-FA*.} We also present Heckman-FA* as alternative option to Heckman-FA, where users extract $\bm{X}^{(p)}$ by simply ranking the selection features based on the largest value of $\hat{\pi}_{2k}$ instead of using Algorithm \ref{alg:evaloptimalassignment}. In other words, we rank the likeliest selection features to be assigned as prediction features based on the objective in Eq. (\ref{eq:optim}). We then examine the first $J$ selection features in the ranking for all $J\in \{1,\ldots,K-1\}$. Letting the first $J$ selection features in the ranking be prediction features, we run the Heckman selection model on the training set and obtain $\underline{\hat{\bm{\beta}}}$. We then compute $\hat{\sigma}^2$, $\hat{\rho}$, and $R^2_a$ using the same steps indicated in lines 5-7 in Algorithm \ref{alg:evaloptimalassignment}. Finally, out of all $J\in \{1,\ldots,K-1\}$, we choose the set of top $J$ selection features in the ranking as prediction features such that $\hat{\rho}\in [\rho_{min}, \rho_{max}]$ and $R^2_a$ value is at a maximum given $\hat{\rho}$. As a result, the columns of $\bm{X}^{(p)}$ corresponding to these features do not equal $\bm{0}$. 

\begin{table*}[t]
    \centering
    \caption{Performance of baselines against Heckman-FA* (Top $J$ Features).}
    \resizebox{0.85\textwidth}{!}{\begin{tabular}{|c|c|c||c|c||c|c||c|c|}
        \hline
        \multirow{2}{*}{Extracted prediction features} & \multirow{2}{*}{$\hat{\rho}$} & \multirow{2}{*}{$R^2_a$} & \multicolumn{2}{c||}{\textbf{Naive}} & \multicolumn{2}{c||}{\textbf{RU}\cite{sahoo2022learning}} & \multicolumn{2}{c|}{\textbf{Heckman-FA*}} \\ \cline{4-9}
        & & & Train MSE & Test MSE & Train MSE & Test MSE & Train MSE & Test MSE \\
        \hline
        \hline
        \multicolumn{9}{|c|}{CRIME} \\
        \hline
        \hline
        2, 3, 8, 12, 16, 18, 21, 24 & 0.0609 & 0.4375 & 0.0172 & 0.0236 & 0.0142 & 0.0232 & 0.0172 & \textbf{0.0216} \\
        \hline
        \hline
        \multicolumn{9}{|c|}{COMPAS} \\
        \hline
        \hline
        2, 5, 6, 10 & 0.3551 & 0.3392 & 0.0769 & 0.2577 & 0.0764 & 0.2590 & 0.0767 & \textbf{0.2495} \\
        \hline
    \end{tabular}}
    \label{tab:topjfeats}
\end{table*}

\noindent \textbf{Computational Complexity.} To derive the computational complexity of Algorithm \ref{alg:heckmanpsi} (Heckman-FA), we first have to consider the complexity of Algorithm \ref{alg:assign}. The assignment computation takes $O(nK)$ time. Because the assignment computation is repeated for $T$ epochs as $\psi$ is trained, the complexity of lines 4-10 is $O(nKT)$. The complexity of Algorithm \ref{alg:evaloptimalassignment} is $O(nKB)$ as the assignment computation is repeated for $B$ Gumbel-Softmax samples. Thus Heckman-FA has a computational complexity of $O(nK(T + B))$.

We also consider the complexity of Heckman-FA*. Similar to Heckman-FA, we first see that $\psi$ is trained in $O(nKT)$ time when running Heckman-FA*. However, the complexity of extraction is different for Heckman-FA* than for Heckman-FA. Since the Heckman model is called for $K-1$ sets of selection features, the extraction process runs in $O(m(K-1))$ time. Thus Heckman-FA* runs in $O(nKT + m(K-1))$ time.

\section{Experiments}

\subsection{Setup}

We evaluate the performance of our framework on the CRIME \cite{adult} and COMPAS \cite{compas} datasets. The CRIME dataset contains socio-economic, law enforcement, and crime information for $1,994$ communities in the United States. For each community, we predict the total number of violent crimes committed (per 100,000 population). The COMPAS dataset consists of $5,278$ records collected from defendants in Florida. We predict each defendant's decile score.

We split each dataset to include 70\% of samples in $\mathcal{D}_{tr}$ and the remaining 30\% as testing samples. We then construct $\mathcal{D}_s$ based on $\mathcal{D}_{tr}$. For the CRIME dataset, we create sample selection bias in $\mathcal{D}_{tr}$ by selecting communities such that the proportion of people under poverty is less than 0.05. As a result, 976 out of 1,395 communities in $\mathcal{D}_{tr}$ are in $\mathcal{D}_s$. For the COMPAS dataset, we select defendants in $\mathcal{D}_{tr}$ who have a violent decile score of less than 10. As a result, there are 2,585 samples in $\mathcal{D}_s$.
We provide the set of selection features used for each dataset in Table \ref{tab:selfeatures} in the Appendix, where $K=26$ for the CRIME dataset and $K=10$ for the COMPAS dataset.

\noindent \textbf{Baselines and Hyperparameters.} We compare our approach to the following baselines: (1) naive linear regression (Naive) on $\mathcal{D}_s$ and (2) Rockafellar-Uryasev (RU) regression \cite{sahoo2022learning}, which is a deep learning approach that involves fitting two neural networks with the RU loss to train a robust model under bounded MAR sample selection bias. Unlike Heckman-FA, the baselines do not have access to selection features. 


As we compute $\hat{\bm{\pi}}$ using Heckman-FA, we work with the following hyperparameters. When initializing $\bm{\pi}$, we set $c=0.75$. We then train $\psi$ for $T=4,000$ epochs with the softmax temperature $\tau=1$ for both datasets. We set the learning rate $\alpha$ equal to 0.75 and 0.05 for the CRIME and COMPAS datasets, respectively. For the extraction of prediction features, we draw $B=1,000$ Gumbel-Softmax samples. In all experiments, the range $[\rho_{min}, \rho_{max}]$ is set to be $[0.01, 0.1]$ and $[0.3, 0.5]$ for the CRIME and COMPAS datasets, respectively. All models are implemented using Pytorch and executed on the Dell XPS 8950 9020 with a Nvidia GeForce RTX 3080 Ti. 

\subsection{Results on Heckman-FA}

In Table \ref{tab:sampling}, we report the training and testing MSE to compare Heckman-FA to the other baselines when using the extracted prediction features. We first observe that the training MSE for Heckman-FA is lower than Naive on the COMPAS dataset yet equal to Naive on the CRIME dataset. This is due to using smaller $\rho_{min}$ and $\rho_{max}$ values for the CRIME dataset. However, the testing MSE for Heckman-FA is 0.0203 and 0.2504 on the CRIME and COMPAS datasets, respectively, which is lower than the testing MSE for Naive. For both datasets, we see that Heckman-FA outperforms Naive given the extracted prediction features. When comparing Heckman-FA and RU, the training MSE of RU is lower than Heckman-FA on the CRIME and COMPAS datasets. This is expected as RU fits a non-linear model for regression. On the other hand, the testing MSE for Heckman-FA is 0.0009 and 0.0027 lower than RU for the CRIME and COMPAS datasets, respectively.  

We also run a paired $t$-test on 10 different prediction feature assignments to analyze the significance of comparing Heckman-FA to the other baselines. Table \ref{tab:significance} in the Appendix shows results of the test. We find that the p-value is very small after running the hypothesis test on both datasets. Given that Heckman-FA significantly outperforms Naive and RU, the results in the two tables show that Heckman-FA outputs a robust regression model under MNAR sample selection bias.

\noindent \textbf{Sensitivity Analysis.} We  perform sensitivity analysis on Heckman-FA by testing the approach over different values for the number of epochs $T$, fixed initial value $c$, and number of Gumbel-Softmax samples $B$ drawn during assignment extraction. Table \ref{tab:testingmsespi0} gives the testing MSE of Heckman-FA across different values of $T$ and $c$ while fixing $B=1,000$. For most combinations of $T$ and $c$ listed in Table \ref{tab:testingmsespi0}, the testing MSEs of Heckman-FA are almost equal to each other for both datasets. We have a similar observation for each combination of $T$ and $B$ as shown in the right three columns of Table \ref{tab:executiontimes} in the Appendix. These results show that Heckman-FA is not sensitive to changes in how $\bm{\pi}$ is initialized and the number of Gumbel-Softmax samples examined during extraction. 


\noindent \textbf{Execution Time.} We report the execution time after running Heckman-FA across different values of $T$ and $B$ in the left three columns of Table \ref{tab:executiontimes} in the Appendix. For both datasets, Heckman-FA runs fast for each combination of $T$ and $B$.

\subsection{Results on Heckman-FA*}

In Table \ref{tab:topjfeats}, we compare Heckman-FA* to the baselines. Under the extracted prediction features, while the training MSE of Heckman-FA* is equal to (lower than) Naive on the CRIME (COMPAS) dataset, the testing MSE for Heckman-FA* is 0.0020 and 0.0082 lower than Naive on the CRIME and COMPAS datasets, respectively. This shows that by simply choosing $J$ prediction features after ranking selection features based on $\hat{\pi}_{2k}$, Heckman-FA* is robust against MNAR sample selection bias. Moreover, in terms of the testing MSE, Heckman-FA* outperforms RU by 0.0016 and 0.0095 on the CRIME and COMPAS datasets, respectively.

\section{Conclusion}

In this paper, we introduced Heckman-FA, a novel data-driven approach that obtains an assignment of prediction features for the Heckman selection model to robustly handle MNAR sample selection bias. Given features that fit the selection of samples, our approach first trains an assignment function by minimizing the MAE on the set of fully observed training samples. Heckman-FA finds prediction features for the Heckman model by drawing a number of Gumbel-Softmax samples using the learned probability of assignment for each selection feature. This set is extracted based on the prediction model's goodness-of-fit and the estimated correlation between noise terms. We observed that Heckman-FA produces a robust regression model under MNAR sample selection bias on the outcome after training the model on real-world datasets. 



\section*{Acknowledgements}
This work was supported in part by NSF 1946391 and 2137335.

\bibliographystyle{plain}
\bibliography{ijcnn}

\begin{thebibliography}{10}

\bibitem{barnighausen2011correcting}
T.~B{\"a}rnighausen, J.~Bor, S.~Wandira-Kazibwe, and D.~Canning.
\newblock Correcting hiv prevalence estimates for survey nonparticipation using heckman-type selection models.
\newblock {\em Epidemiology}, 2011.

\bibitem{bushway2007magic}
S.~Bushway, B.~D. Johnson, and L.~A. Slocum.
\newblock Is the magic still there? the use of the heckman two-step correction for selection bias in criminology.
\newblock {\em Journal of quantitative criminology}, 23:151--178, 2007.

\bibitem{chen2016robust}
X.~Chen, M.~Monfort, A.~Liu, and B.~D. Ziebart.
\newblock Robust covariate shift regression.
\newblock In {\em AISTATS}, 2016.

\bibitem{du2022fair}
W.~Du, X.~Wu, and H.~Tong.
\newblock Fair regression under sample selection bias.
\newblock In {\em IEEE Big Data}, 2022.

\bibitem{adult}
D.~Dua and C.~Graff.
\newblock {UCI} machine learning repository.
\newblock \url{http://archive.ics.uci.edu/ml}, 2017.

\bibitem{genback2015uncertainty}
M.~Genb{\"a}ck, E.~Stanghellini, and X.~de~Luna.
\newblock Uncertainty intervals for regression parameters with non-ignorable missingness in the outcome.
\newblock {\em Statistical Papers}, 56:829--847, 2015.

\bibitem{heckman1979sample}
J.~J. Heckman.
\newblock Sample selection bias as a specification error.
\newblock {\em J. Econom.}, pages 153--161, 1979.

\bibitem{hu2022non}
X.~Hu, Y.~Niu, C.~Miao, X.~S. Hua, and H.~Zhang.
\newblock On non-random missing labels in semi-supervised learning.
\newblock {\em arXiv:2206.14923}, 2022.

\bibitem{jang2016categorical}
E.~Jang, S.~Gu, and B.~Poole.
\newblock Categorical reparameterization with gumbel-softmax.
\newblock {\em arXiv:1611.01144}, 2016.

\bibitem{kahng2023domain}
H.~Kahng, H.~Do, and J.~Zhong.
\newblock Domain generalization via heckman-type selection models.
\newblock In {\em ICLR}, 2023.

\bibitem{compas}
J.~Larson, S.~Mattu, L.~Kirchner, and J.~Angwin.
\newblock Compas dataset.
\newblock \url{https://github.com/propublica/compas-analysis/}, 2017.

\bibitem{lee2021dual}
J.~W. Lee, S.~Park, and J.~Lee.
\newblock Dual unbiased recommender learning for implicit feedback.
\newblock In {\em SIGIR}, 2021.

\bibitem{lei2021near}
Q.~Lei, W.~Hu, and J.~Lee.
\newblock Near-optimal linear regression under distribution shift.
\newblock In {\em ICML}, 2021.

\bibitem{leung1996choice}
S.~F. Leung and S.~Yu.
\newblock On the choice between sample selection and two-part models.
\newblock {\em Journal of Econometrics}, 72(1-2):197--229, 1996.

\bibitem{lippe2022citris}
P.~Lippe, S.~Magliacane, S.~L{\"o}we, Y.~M. Asano, T.~Cohen, and S.~Gavves.
\newblock Citris: Causal identifiability from temporal intervened sequences.
\newblock In {\em ICML}, 2022.

\bibitem{moreno2012unifying}
J.~G. Moreno-Torres, T.~Raeder, R.~Alaiz-Rodr{\'\i}guez, N.~V. Chawla, and F.~Herrera.
\newblock A unifying view on dataset shift in classification.
\newblock {\em Pattern recognition}, 45(1):521--530, 2012.

\bibitem{nawata1993note}
K.~Nawata.
\newblock A note on the estimation of models with sample-selection biases.
\newblock {\em Econ. Lett.}, 42(1):15--24, 1993.

\bibitem{puhani2000heckman}
P.~Puhani.
\newblock The heckman correction for sample selection and its critique.
\newblock {\em J. Econ. Surv.}, 14(1):53--68, 2000.

\bibitem{sahoo2022learning}
R.~Sahoo, L.~Lei, and S.~Wager.
\newblock Learning from a biased sample.
\newblock {\em arXiv:2209.01754}, 2022.

\bibitem{shimodaira2000improving}
H.~Shimodaira.
\newblock Improving predictive inference under covariate shift by weighting the log-likelihood function.
\newblock {\em J. Stat. Plan. Inference}, 90(2):227--244, 2000.

\bibitem{vella1998estimating}
F.~Vella.
\newblock Estimating models with sample selection bias: a survey.
\newblock {\em Journal of Human Resources}, pages 127--169, 1998.

\bibitem{wang2019doubly}
X.~Wang, R.~Zhang, Y.~Sun, and J.~Qi.
\newblock Doubly robust joint learning for recommendation on data missing not at random.
\newblock In {\em ICML}, 2019.

\bibitem{wolfolds2019misaccounting}
S.~E. Wolfolds and J.~Siegel.
\newblock Misaccounting for endogeneity: The peril of relying on the heckman two-step method without a valid instrument.
\newblock {\em Strategic Management Journal}, 40(3):432--462, 2019.

\end{thebibliography}

\clearpage

\appendix

\begin{table}[t]
    \small
    \centering
    \caption{Description of main symbols used in paper.}
    \resizebox{0.47\textwidth}{!}{\begin{tabular}{|c|l|}
    \hline
        \textbf{Notation} & \textbf{Description} \\
    \hline
        $m (n)$ & size of $\mathcal{D}_s (\mathcal{D}_{tr})$ \\
        $\mathcal{D}_s$ ($\mathcal{D}_u$) & training samples with observed (missing) outcome \\
        $\bm{x}_{i}^{(s)}$ ($\bm{x}_{i}^{(p)}$), $y_i$ & selection (prediction) features, outcome of a sample \\
        $s_i$ & selection index \\
        $K$ ($J$) & number of selection (prediction) features \\
        $\rho$ & correlation coefficient between the noise terms \\
        $\psi$ & assignment function \\
        $\bm{Z}$ & categorical sample matrix \\
        $\pi_{2k} (\pi_{1k})$ & probability of (not) assigning $k$th selection feature \\
        $T$ & number of training epochs \\
        $B$ & number of Gumbel-Softmax samples (extraction) \\
    \hline
    \end{tabular}}
    \label{tab:notation}
\end{table}


\noindent \textbf{Discussion on Training Assignment Function by Minimizing MSE or RMSE.} Although other metrics such as mean squared error (MSE) and root mean squared error (RMSE) are typically used to evaluate the performance of linear regression, we cannot minimize these metrics in order to obtain $\hat{\bm{\pi}}$. We show that if our objective is to minimize the MSE or RMSE, then $\bm{\pi}$ does not change at all during backpropagation, i.e. $\nabla_{\bm{\pi}}\mathcal{L}$ is a zero matrix. As an example, consider $\mathcal{L}_{MSE}$, which denotes the MSE loss function. For $\nabla_{\bm{\pi}} \mathcal{L}_{MSE}$, we see that 
\begin{equation} \small
    \begin{split}
        \frac{ \partial \mathcal{L}_{MSE} }{ \partial \pi_{qk} } &= \frac{ \partial \mathcal{L}_{MSE} }{ \partial\hat{y}_i } \cdot \frac{ \partial\hat{y}_i }{ \partial x^{(p)}_{ik} } \cdot \frac{ \partial x^{(p)}_{ik} }{ \partial \psi(k) } \cdot \frac{ \partial \psi(k) }{ \partial\tilde{z}_{qk} }\cdot \frac{ \partial\tilde{z}_{qk} }{ \partial \pi_{qk} } \\
            &=  \frac{ \partial \mathcal{L}_{MSE} }{ \partial \psi(k) } \cdot \frac{ \partial \psi(k) }{ \partial\tilde{z}_{qk} }\cdot \frac{ \partial\tilde{z}_{qk} }{ \partial \pi_{qk} } \\
    \end{split}
\end{equation}
where
\begin{equation} \small
    \begin{split}
        \frac{ \partial \mathcal{L}_{MSE} }{ \partial \psi(k) } &= -\frac{2}{m} \sum_{i=1}^m (y_i - \hat{y}_i) \cdot \beta_{k} \cdot x^{(s)}_{ik} \\
            &= -\frac{2}{m}\beta_{k}\sum_{i=1}^m (y_i - \hat{y}_i)\cdot x^{(s)}_{ik} \\
            &= -\frac{2}{m}\beta_{k}\sum_{i=1}^m v_i\cdot x^{(s)}_{ik}
    \end{split}
\end{equation}
where $v_i$ is the error of predicting the outcome of $\bm{t}_i\in \mathcal{D}_s$ using the Heckman selection model. Given that $\bm{\beta}$ has length $K$, consider two cases on the $k$th selection feature. First, for any $k$th selection feature not assigned for prediction, $\bm{X}^{(p)}_{\cdot k} = \bm{0}$, where $\bm{X}^{(p)}_{\cdot k}$ is the $k$th column of $\bm{X}^{(p)}$. Thus $\beta_k = 0$ after running the Heckman selection model and hence $\displaystyle \frac{ \partial \mathcal{L}_{MSE} }{ \partial \psi(k) } = 0$. Second, for any $k$th selection feature assigned for prediction, $x^{(s)}_{ik} = x^{(p)}_{ik}$ for all $\bm{t}_i\in \mathcal{D}_s$. Thus
\begin{equation} \small
    \begin{split}
        \frac{ \partial \mathcal{L}_{MSE} }{ \partial \psi(k) } = -\frac{2}{m}\beta_{k}\sum_{i=1}^m v_i\cdot x^{(s)}_{ik} &= -\frac{2}{m}\beta_{k}\sum_{i=1}^m v_i\cdot x^{(p)}_{ik} \\
            &= -\frac{2}{m}\beta_{k}\left( \bm{v}^{T}\bm{X}^{(p)}_{\cdot k} \right) \\
    \end{split}
\end{equation}
where $\bm{v}$ is the error vector. Now $\bm{X}^{(p)}_{\cdot k}$ is considered part of the input space that is used to compute $\hat{\bm{y}}$ and in turn $\bm{v}$. Because the values in $\bm{\beta}$ are estimated by minimizing $\bm{v}^{T}\bm{v}$, then $\bm{v}$ is orthogonal to the columns of the input space used to compute $\hat{\bm{y}}$, which includes $\bm{X}^{(p)}_{\cdot k}$. Thus $\bm{v}^{T} \bm{X}^{(p)}_{\cdot k}=0$, and $\displaystyle \frac{ \partial \mathcal{L}_{MSE} }{ \partial \psi(k) } = 0$ for all selection features assigned for prediction. Based on these two cases, $\displaystyle \frac{ \partial \mathcal{L}_{MSE} }{ \partial \pi_{qk} } = 0$ for all $K$ selection features. Hence $\nabla_{\bm{\pi}}\mathcal{L}_{MSE}$ is a zero matrix.

\begin{table}[t]
    \small
    \centering
    \caption{Features used for selection.}
    \resizebox{0.5\textwidth}{!}{\begin{tabular}{|l|l|}
    \hline
        \textbf{Dataset} & Selection Features \\
    \hline
        CRIME & householdsize, racepctblack, racePctWhite, \\
            & racePctAsian, racePctHisp, agePct12t21, agePct12t29, \\
            &  agePct16t24, agePct65up, numbUrban, pctUrban, \\
            &   medIncome, pctWWage, pctWFarmSelf, pctWInvInc, \\
            &   pctWSocSec, pctWPubAsst, pctWRetire, medFamInc, \\
            &   perCapInc, whitePerCap, HispPerCap, blackPerCap, \\
            &  indianPerCap, AsianPerCap, OtherPerCap \\
    \hline
        COMPAS & sex, age, juv\_fel\_count, juv\_misd\_count, \\
            & priors\_count, two\_year\_recid, age\_cat\_25 - 45, \\
            & age\_cat\_Greater than 45, race\_Caucasian, \\
            & c\_charge\_degree\_M \\
    \hline
    \end{tabular}}
    \label{tab:selfeatures}
\end{table}


\begin{table}[t]
    \centering
    \caption{Results of paired $t$-test on the difference between the testing MSE of Heckman-FA and other baselines.}
    \resizebox{0.45\textwidth}{!}{\begin{tabular}{|c||c|c|}
    \hline
        Comparison & Test MSE Difference & P-value \\
    \hline
    \hline
    \multicolumn{3}{|c|}{CRIME} \\
    \hline
    \hline
       Heckman-FA vs. Naive  & -0.0008 $\pm$ 0.0004 & 0.0001 \\
       Heckman-FA vs. RU  & -0.0007 $\pm$ 0.0007 & 0.0053 \\
    \hline
    \hline
    \multicolumn{3}{|c|}{COMPAS} \\
    \hline
    \hline
      Heckman-FA vs. Naive  & -0.0068 $\pm$ 0.0006 & 7.723e-11 \\
      Heckman-FA vs. RU  & -0.0059 $\pm$ 0.0051 & 0.0055 \\
    \hline
    \end{tabular}}
    \label{tab:significance}
\end{table}

\begin{table}[t]
    \centering
    \caption{Execution times (in seconds) and testing MSEs of Heckman-FA using different values of $T$ and $B$.}
    \resizebox{0.45\textwidth}{!}{\begin{tabular}{|c||c|c|c||c|c|c|}
        \hline
        \multirow{3}{*}{$B$} & \multicolumn{6}{c|}{$T$ training epochs}  \\ \cline{2-7}
        & \multicolumn{3}{c||}{Execution time} & \multicolumn{3}{c|}{Testing MSE} \\ \cline{2-7}
        & 100 & 500 & 1,000 & 100 & 500 & 1,000 \\
        \hline
        \hline
        \multicolumn{7}{|c|}{CRIME} \\
        \hline
        \hline
        100 & 5.32 & 17.67 & 29.33 & 0.0209 & 0.0206 & 0.0210 \\
        500 & 16.02 & 26.06 & 44.23 & 0.0213 & 0.0210 & 0.0209 \\
        1,000 & 37.84 & 56.87 & 75.28 & 0.0212 & 0.0206 & 0.0209 \\
        \hline
        \hline
        \multicolumn{7}{|c|}{COMPAS} \\
        \hline
        \hline
        100  & 7.70 & 21.30 & 39.32 & 0.2503 & 0.2502 & 0.2505 \\
        500 & 22.16 & 37.39 & 54.05 & 0.2506 & 0.2506 & 0.2503 \\
        1,000 & 49.72 & 70.46 & 97.64 & 0.2502 & 0.2506 & 0.2502 \\
        \hline
    \end{tabular}}
    \label{tab:executiontimes}
\end{table}

\noindent \textbf{Execution Time.} We report the execution time after running Heckman-FA across different values of $T$ and the number of Gumbel-Softmax samples $B$ drawn during the assignment extraction in left three columns of Table \ref{tab:executiontimes}. We observe that for both datasets, Heckman-FA runs fast for each combination of $T$ and $B$. For instance, when $T=100$ and $B=100$, Heckman-FA is completed after 5.32 and 7.70 seconds for the CRIME and COMPAS datasets, respectively. While the execution time increases as $T$ and $B$ increase, the testing MSE of Heckman-FA remains close to the testing MSE reported in Table \ref{tab:sampling}. This result shows that Heckman-FA is fast while maintaining quality performance on the testing set.

\begin{table}[t] 
    \centering
    \caption{Testing MSEs of Heckman-C and Heckman-FA* on the CRIME dataset across different values of $J$ for the number of prediction features based on top $J$ selection features. Underlined testing MSE corresponds to extracted set.}
        \resizebox{0.3\textwidth}{!}{\begin{tabular}{|c||c||c|}
            \hline
            $J$ & Heckman-C & Heckman-FA* \\
            \hline
            6 & 0.0490 & 0.0290 \\
            7 & 0.0563 & 0.0290 \\
            8 & 0.0479 & \underline{0.0216} \\
            9 & 0.0500 & 0.0215 \\
            10 & 0.0498 & 0.0216 \\
            11 & 0.0555 & 0.0216 \\
            12 & 0.0524 & 0.0216 \\
            \hline
        \end{tabular}}
    \label{tab:moretopj}
\end{table}

\noindent \textbf{Comparison of Heckman-FA* with Correlation-Based Ranking.} To further show the effectiveness of Heckman-FA*, which requires training $\psi$, we also compare the ranking of selection features based on $\hat{\pi}_{2k}$ to the ranking of selection features based on their strength of correlation with the outcome. We use the name Heckman-C to describe the correlation-based ranking of selection features. Unlike the ranking based on $\hat{\pi}_{2k}$, Heckman-C does not rely on training the assignment function beforehand. For Heckman-FA* and Heckman-C, we run each approach on the CRIME dataset using the top $J$ selection features across different values of $J$.
Table \ref{tab:moretopj} provides the MSE of the models on the testing set. We consider the values of $J=6$ through $J=12$. The underlined testing MSE corresponds to the size of the final set of prediction features. For Heckman-FA*, the final set of prediction features consists of $8$ prediction features. For Heckman-C, there is no testing MSE underlined as there is no set of top $J$ selection features extracted for the final set of prediction features. In other words, when ranking the selection features based on their correlation with the outcome using the CRIME dataset, there is no set of $J$ features from the ranking that ensures the robustness of Heckman-C under MNAR sample selection bias. In our experiment, we also find that the condition $\hat{\rho}\in [\rho_{min}, \rho_{max}]$ is satisfied for Heckman-FA* for the range $J=8$ through $J=12$. However, for Heckman-C, no set of $J$ prediction features satisfies this condition for any value of $J$. This indicates that when assigning prediction features for the Heckman model, ranking selection features based on $\hat{\pi}_{2k}$ after training $\psi$ is more effective than ranking based on the correlation of features with the outcome.

\end{document}